# Universal Spectral Transfer with Physical Prior-Informed Deep Generative Learning

Yanmin Zhu[1], Loza F. Tadesse[1,2,3,*]
[1]Department of Mechanical Engineering, MIT, Cambridge, MA, USA
[2]Ragon Institute of MGH, MIT and Harvard, Cambridge, MA, USA
[3]Jameel Clinic for AI & Healthcare, MIT, Cambridge, MA, USA

**Abstract**
Spectroscopy is a powerful analytical technique for characterizing matter across physical and biological realms[1-5]. However, its fundamental principle necessitates specialized instrumentation per physical phenomena probed, limiting broad adoption and use in all relevant research. In this study, we introduce SpectroGen, a novel physical prior-informed deep generative model for generating relevant spectral signatures across modalities using experimentally collected spectral input only from a single modality. We achieve this by reimagining the representation of spectral data as mathematical constructs of distributions instead of their traditional physical and molecular state representations. The results from 319 standard mineral samples tested demonstrate generating with 99% correlation and 0.01 root mean square error with superior resolution than experimentally acquired ground truth spectra. We showed transferring capability across Raman, Infrared, and X-ray Diffraction modalities with Gaussian, Lorentzian, and Voigt distribution priors respectively[6-10]. This approach however is globally generalizable for any spectral input that can be represented by a distribution prior, making it universally applicable. We believe our work revolutionizes the application sphere of spectroscopy, which has traditionally been limited by access to the required sophisticated and often expensive equipment towards accelerating material, pharmaceutical, and biological discoveries.

## Main

Material characterization aims to elucidate the microstructure, composition, and other intrinsic properties of materials[11]. This process assists in optimizing and evaluating material performance[12,13], designing novel materials, understanding material behavior in ecological environments[14-17], and enhancing energy conversion efficiency in energy materials[18,19]. Material characterization can be precisely performed using advanced microscopy and spectroscopy techniques[20,21]. Infrared (IR) absorption unveils functional groups within molecules[22], Raman scattering generates information on molecular vibrations, symmetry and crystal structures[23], X-ray scattering determines the elemental composition[24], while X-ray diffraction (XRD) visualizes crystal structures[25]. Spectroscopy based on ultraviolet-visible range elucidates conjugated systems within molecules using absorption peaks arising from electronic transitions[26]. It is worth noting that specific equipment is needed for each type of interrogation due to the distinct physical principles underpinning each spectroscopic approach. Computationally generating accurate models of these distinct molecular and crystallographic material dynamics using their canonical understanding is an almost impossible task given the complexity and computational load.

Deep learning techniques for material characterization encompass performance prediction[27], the analysis and preprocessing of microscopic images and spectroscopic data[28-31], and the design of novel materials[32]. Distinct from conventional deep learning techniques, which primarily focus on classification and regression predictions, generative artificial intelligence (AI) techniques emphasize understanding the underlying structure and features of data. Generative AI has demonstrated exceptional capabilities across various biological and physical domains, successfully applied in areas such as gene editing[33] and protein design[34]. Nevertheless, its application in material characterization remains in its nascent stages. In this study, we reimagine spectra as a mathematical distribution construct instead of its typical physical representation of atomic and molecular interaction in matter. By providing these distributions as a physical-prior input we enable a generative AI model, SpectroGen, that can successfully generate any spectra type using only experimentally generated input from a single spectroscopy modality. SpectroGen generates spectra of superior resolution and with 99% correlation to experimentally acquired ground truth spectra. Our method revolutionizes the generation of spectroscopic data, offering precise, efficient, and rapid tools for material characterization. It serves as a powerful asset in the design and discovery of novel materials.

## Cross-Domain Spectral Transfer via Generation

SpectroGen transforms spectra by leveraging priors of spectral broadening distributions of vibrations and absorptions, coupled with a deep generative model that specializes on tracking curves. To implement, first we establish a probabilistic encoder $q_\phi(z|x)$ that learns the prior probability distribution of experimentally derived input Spectrum A, capturing the physical constraints inherent in the spectral transformation process, including the complex dependencies of line broadening, superposition, and wavenumber shifts. The Variational Autoencoder (VAE), implemented in our work as the frame of encoder and decoder, has demonstrated its efficacy in learning high-dimensional complex probability distributions across various domains, such as predicting gene mutations by capturing genetic variants[35] and facilitating the heuristic design of novel therapeutic drug molecules[36]. The algorithm training occurs through multiple waveform distribution analyses to deconstruct single-frequency peaks, broadening, wavenumber shifts, and deformations of composite superimposed waveforms (Fig.1a). The intermediate extracted features are captured in latent low-dimensional vectors $z$. A probabilistic decoder $p_\theta(z)$ then up-samples and reconstructs the probability distribution of the generated Spectrum B from the learned approximate posterior distribution. The stability and performance of the spectral generation are verified by thorough physical prior spectral deconstruction (Fig.1b), such as Gaussian distribution prior, Voigt distribution prior, and Lorentzian distribution prior, and model fitting (Fig.1c and d).

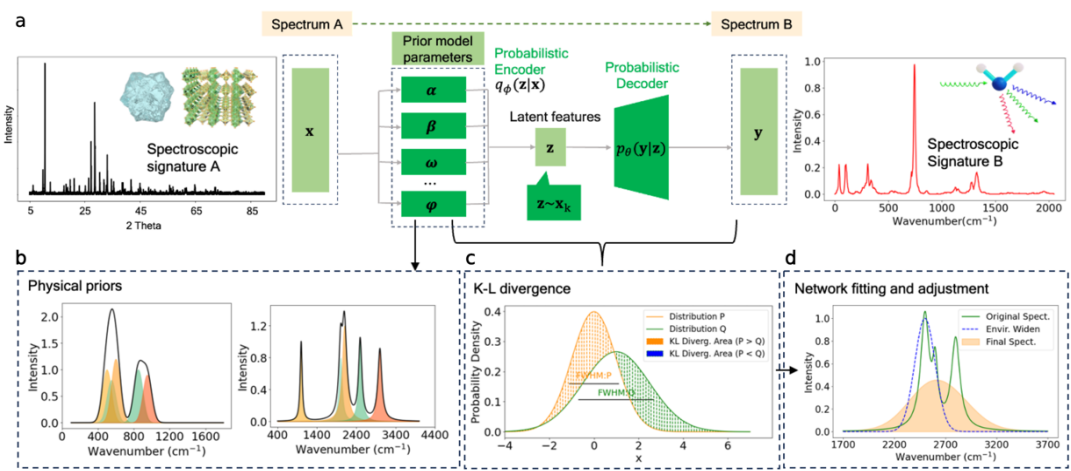

Fig.1 Modeling strategy. (a) SpectroGen employs an experimentally derived spectrum from modality A. (b) Physical priors are employed to deconstruct the spectral distribution and map the spectral distribution to latent features $z$ via a frequency encoder $q_\phi(z|x)$ (see Supplemental method Sec.3). SpectroGen designs a probabilistic decoder $p_\theta(y|z)$ to reconstruct the second (generated) spectral distribution. (c) SpectroGen computes Kullback-Leibler (K-L) divergence to perform spectral fitting (see Supplemental method Sec.5). (d) The network's fitting capabilities are fully exploited to accommodate the irregular environmental broadening present in the spectra (see Supplemental method Sec.4).

SpectroGen is able to make accurate cross modality spectral generation, primarily because of two key aspects. Firstly, the accurate physical priors that were inputted to represent respective spectra from the modalities of interest (Fig.1b), which removes the original model formulation constraints of the decoder and secondly, the VAE backbone architecture used which is best suited for matching curves. Gaussian distribution and Voigt distribution priors[6-10] were used to fit Infrared and X-ray Diffraction spectra respectively as established by prior works (Supplemental method Sec.3). Kullback-Leibler (KL) divergence loss (Fig.1c, Supplemental method Sec.5) between the generated and input spectra is computed and iteratively minimized during the training phase towards accurate spectral generation[26]. SpectroGen extensively leverages the network's automatic fitting capabilities to accurately address the non-uniform broadening of spectrum (Fig.1d). This approach mitigates the limitations of fitting based solely on physical priors, thereby enabling more precise model transformations.

We demonstrate SpectroGen on RRUFF dataset[38] (Fig.2b and 2g), comprising 6,006 IMF-approved standard mineral samples, from which 319 IR-Raman and 371 XRD-Raman data pairs were examined (see Supplemental data S1-S4). We computed wavenumber shifts, peak heights, and the full width at half maximum (FWHM) of peaks to evaluate the accuracy of spectrum generation (Fig.2c and 2h). Furthermore, we conducted a material source classification task to compare the classification accuracy of the generated spectra with that of experimentally collected spectra, thereby assessing the informational efficacy of the generated spectra (Fig.3, Supplemental Notes 1-2, Supplemental Data 5-6). SpectroGen exhibited 99% correlation on peak characteristics (Fig.2d and 2i) and an average classification accuracy of 94%, surpassing 75% accuracy obtained from experimentally acquired Raman spectra (Fig.3c).

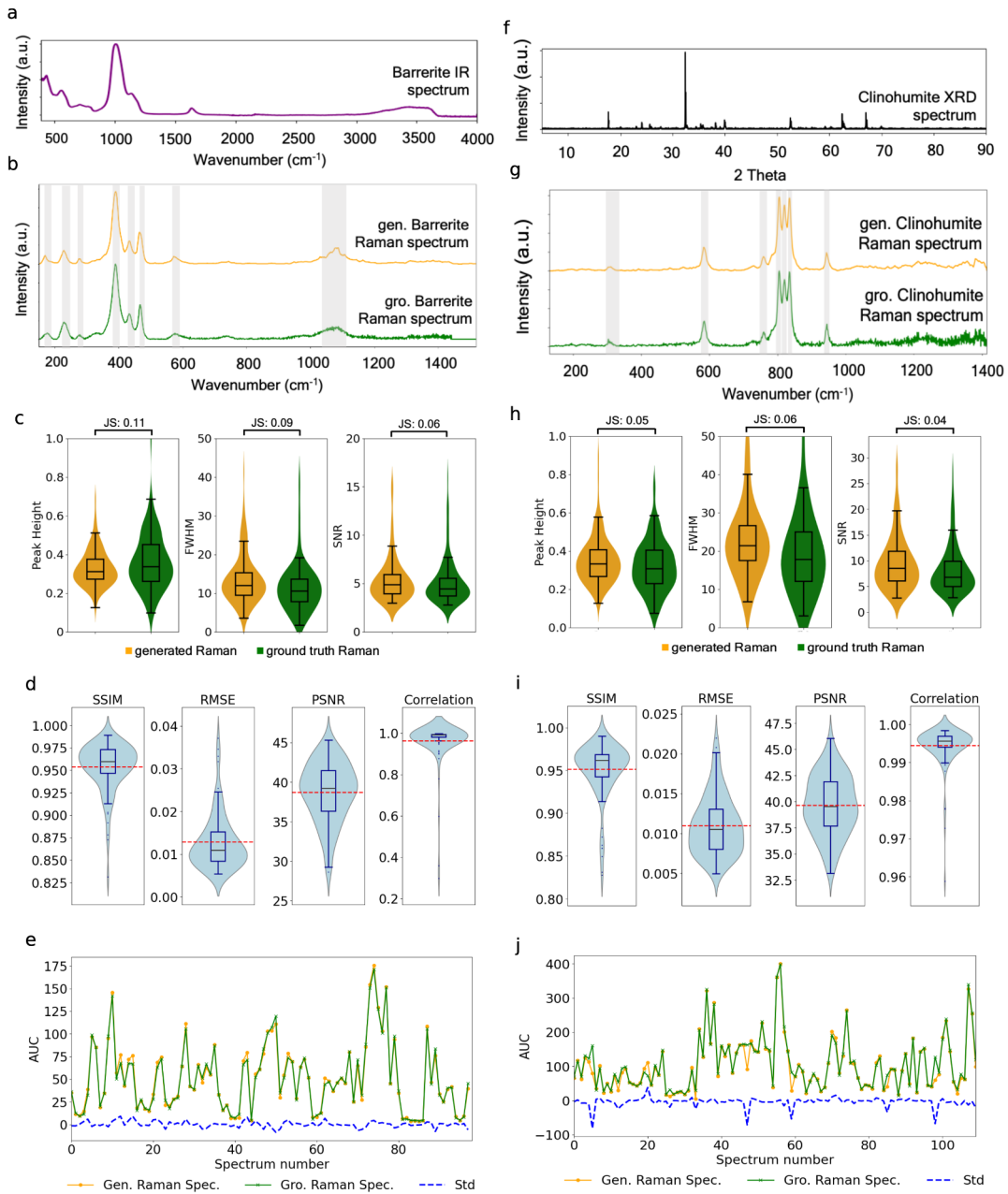

Fig.2. **SpectroGen** accurately generated spectra across modalities. Demonstration of IR to Raman transfer test with (a) Barrerite IR spectrum, (b) generated (yellow) and experimentally collected (green, termed as ground truth) Barrerite Raman spectrum, showing alignment in peak positions and lower noise. Spectral characteristic assessments for the entire generated datasets with (c) average peak height, FWHM and SNR of generated Raman spectra (yellow) and experimentally collected Raman spectra (green). Jensen-Shannon (JS) divergence is calculated to assess the similarity between the generated and experimentally collected spectra. (d) SSIM, RMSE, PSNR and correlation test between generated Raman spectra and experimentally collected Raman spectra. SpectroGen also gives high similarities in image-based assessments. (e) AUC test of generated (Gen.) and experimentally collected (Gro.) Raman spectra and their standard deviation show near 0 deviations. Demonstration of XRD to Raman transfer test with (f) Clinohumite input XRD spectrum, (g) generated and experimentally collected Clinohumite Raman spectrum. SpectroGen accurately predicts the Raman spectrum from the XRD input with a Voigt distribution prior, correctly identifying peak locations with lower noise levels. (h) Average peak height, FWHM and SNR assessments for generated (yellow) and experimentally collected (green) Raman spectra dataset. JS divergences close to zero confirm the strong alignment between them. (i) SSIM, RMSE, PSNR and correlation test between generated Raman spectra and experimentally collected Raman spectra showing SpectroGen has excellent prediction performances on XRD to Raman task. (j) AUC test of generated and ground truth spectra shows alignments between them.

**SpectroGen is as accurate as experimental collection**

The experimentally acquired spectra, aside from conforming to the physical prior, are also influenced by various broadening mechanisms such as collision broadening, Doppler broadening, and transit-time broadening (See Supplemental method 4). A key feature of our method is its ability to fit the difference between physical priors and actual spectra through the network. This allows for accurate fitting of peak overlap and broadening. We present two sets of results each with Gaussian distribution prior for the input of IR (Fig.2a and 2b) (see Supplemental methods eq. s3-1) and Voigt distribution prior for XRD spectra, respectively (Fig.2f and 2g, see Supplemental methods eq. s3-3 and complete generated spectral dataset available in the Supplemental Data 1 and 20). It is well-established that the IR spectra of the solid mineral materials used for experimental data collection conform to a Gaussian distribution prior[9] and the X-ray spectra peaks follow a Voigt distribution[10]. For the transformation from IR to Raman with a Gaussian distribution prior, SpectroGen precisely captures and reconstructs the information of 8 peaks in the Barrerite Raman spectrum (Fig.2b) from their respective IR spectra (Fig.2a). Notably, the generated Raman spectrum matches the respective broadening and peak height and exhibits a smoother waveform. Voigt distribution, which is a convolution of Gaussian and Lorentzian distributions, is used to represent XRD spectra peaks. Similar to the IR-Raman generation task, the generated spectrum precisely matches peak positions, heights, and FWHM (Fig.2h) with lower noise, where the SNR for the generated and experimentally collected Raman spectrum are 11.10 and 3.11, respectively.

Quality of generated spectrum was assessed using spectrum-based and image-based evaluation metrics to assess the performance of the generated spectra, as illustrated in Fig.2. We quantified average peak height, FWHM, and SNR for all the 97 and 110 test pairs of SpectroGen generated and experimentally obtained Raman spectra in the IR-Raman and XRD-Raman dataset respectively. As shown in Fig.2c, 2h, and Table S1, the average peak height and FWHM of the generated spectra exhibit a similar distribution to that of the collected spectra, with Jensen-Shannon (JS) divergences of 0.11 and 0.09 for IR-to-Raman, and 0.05 and 0.06 for XRD-to-Raman, respectively. Notably, the generated spectra on average possess a higher SNR compared to the experimentally collected spectra, which is consistent with the spectral examples provided in Fig.2b and 2g. Structural Similarity Index (SSIM), Root Mean Square Error (RMSE), Peak Signal-to-Noise Ratio (PSNR), and correlation assessments were performed as part of the image-based evaluation, which compares graphical structure and image content. As depicted in Fig.3d and Table S2, for IR to Raman task, SpectroGen generated spectrum has mean SSIM of $0.96 \pm 0.03$, RMSE of $0.010 \pm 0.006$, correlation of $0.99 \pm 0.01$ and PSNR of $39 \pm 4$ dB. These results demonstrate the exceptional similarity of the visually observed trajectory of the generated and experimentally obtained spectrum. In addition, area under the curve (AUC) calculations show good alignments between the generated and ground truth spectra (see Fig.2e are available in Supplemental Data 5-9). Furthermore, an outstanding performance is also demonstrated in the XRD-to-Raman transformation task. As shown in Fig.2i, SpectroGen performs $0.97 \pm 0.04$ mean SSIM, $0.010 \pm 0.009$ RMSE, $43 \pm 4$ dB PSNR, and $0.98 \pm 0.08$ correlation. In the AUC test (see Fig.2j), generated spectra are well aligned with experimentally collected spectra.

## Assessment of Information Transfer Effectiveness via Classification Testing

To determine SpectroGen's information transfer effectiveness, we compared performance on material type classification task using generated spectra and experimentally obtained spectra separately. As shown in Fig.3, generated spectra achieved an average accuracy of 94% across 26 categories of mineral materials (test set: 59%) for 10 rounds of repetitive classification tasks. Under identical network parameter conditions, the experimentally collected spectra had an average classification accuracy of 75% (test set: 67%). This effectively demonstrates not only its effectiveness in transferring the fingerprint information that depicts molecular vibration, but the higher SNR result it generates while maintaining transfer precision. The confusion matrix reveals that the classification performance of generated spectra and experimentally collected spectra is similar for individual categories on a randomly selected training round. (Detailed data from the 10 rounds of repetitive classification tests are available in the supplemental data 39-68). Even though it is beyond the scope of our current study, we generally observe poor classification performance due to the limited number of samples in the dataset; the majority of categories have fewer than five samples. We believe that the slightly lower accuracy observed in the test set of the generated spectra, compared to the experimentally collected spectra, may be attributed to the instability in classification performance resulting from the small dataset size, of much less than 10 spectra per material type. We expect this to improve significantly with a more substantial dataset.

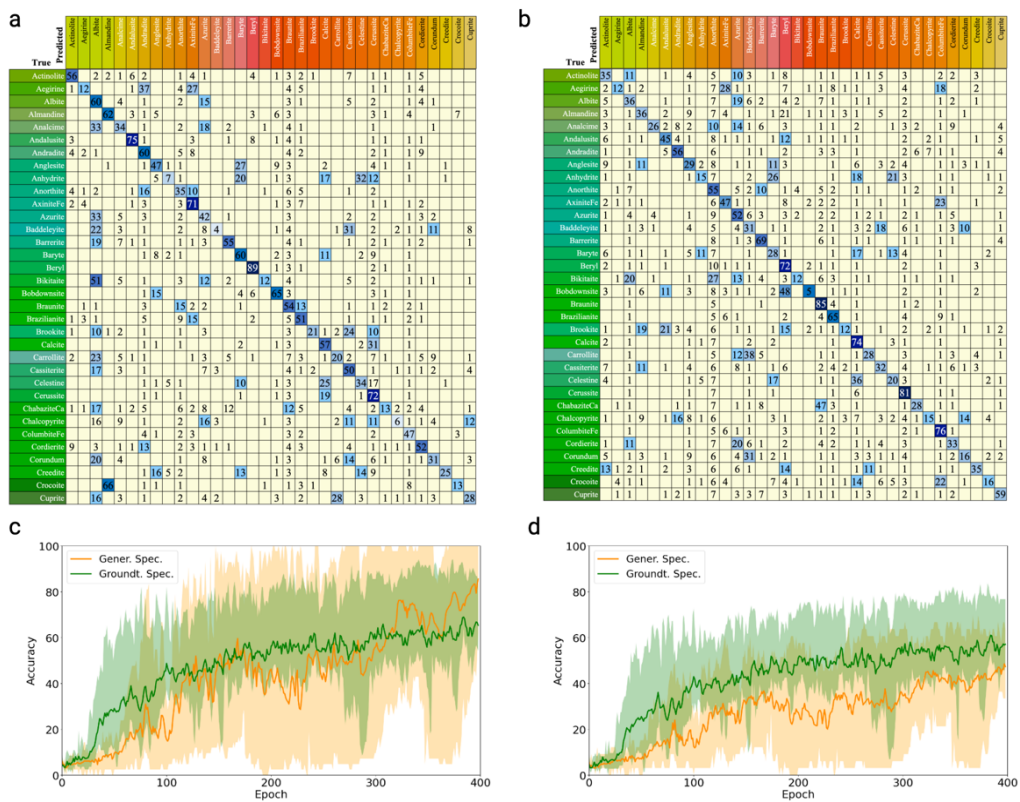

Fig.3 SpectroGen precisely transfers information where generated spectra outperform experimental spectra. (a) Confusion matrix of the classification test using generated Raman spectra. (b) Confusion matrix based on the experimentally collected spectra. The generated spectra deliver a similar classification result with the experimentally collected spectra. (c) Accuracy results based on the train set. Green line: ground truth spectra, orange line: generated spectra. Generated Raman spectra achieve repetitive accuracy results compared to ground truth spectra while reaching a higher final accuracy. (d) Accuracy based on the test set. Generated spectrum with SpectroGen provides competitive information effectiveness as the experimentally collected data.

**SpectroGen interpretability test via physical prior distribution analysis**
We validated the importance of the physical prior in the network by intentionally misrepresenting respective spectra and their distribution. When using a Lorentzian distribution as the physical prior for IR, we obtained an average peak height of 0.59, an average FWHM of 134.54, and an SNR of 47.69 for the generated Raman spectra, compared to an average peak height of 0.39, an average FWHM of 14.75, and an SNR of 5.22 for the experimentally collected Raman spectra. When XRD is incorrectly represented using a Gaussian distribution, the performance of SpectroGen on the generated Raman spectra similarly declines, yielding an average peak height of 0.27, an FWHM of 26.17, and an SNR of 12.87 compared to an average peak height of 0.24, an FWHM of 20.21, and an SNR of 7.88 for the experimentally collected Raman spectra. Similar drops also appear in image-based assessments (See Supplemental Fig.S3, S4, S9 and S10). These results underscore the critical role of physical prior models in the interpretability of the network leading to precise generation, in contrast to a purely black-box approach, which relies solely on the network without incorporating physical priors. To further elucidate the guidance and impact of the physical prior model on the network, we visualized the latent vector values under the three previously tested conditions(Fig.4(a)). The results after 120 training epochs are shown in Fig.4(b) and (c). We observed that different physical priors influenced the latent space in terms of their spatial distributions and values. We further performed principal component analysis (PCA) and calculated the cosine similarity between two distributions, presented in Fig.4(d) and (e). We observed significant differences in the spatial distribution of latent features when using Gaussian versus Lorentzian distribution priors. The Gaussian prior led to a dispersed distribution in principal components 1 and 2 for the IR-to-Raman transformation, while the Lorentzian prior resulted in a concentrated distribution with the same input data. This substitution not only changed the magnitude of latent feature values but also their distribution, affecting the generated results. Cosine similarity calculations revealed differences in the range of [-0.15, 0.20], indicating significant variations in both the magnitude and direction of the latent features.

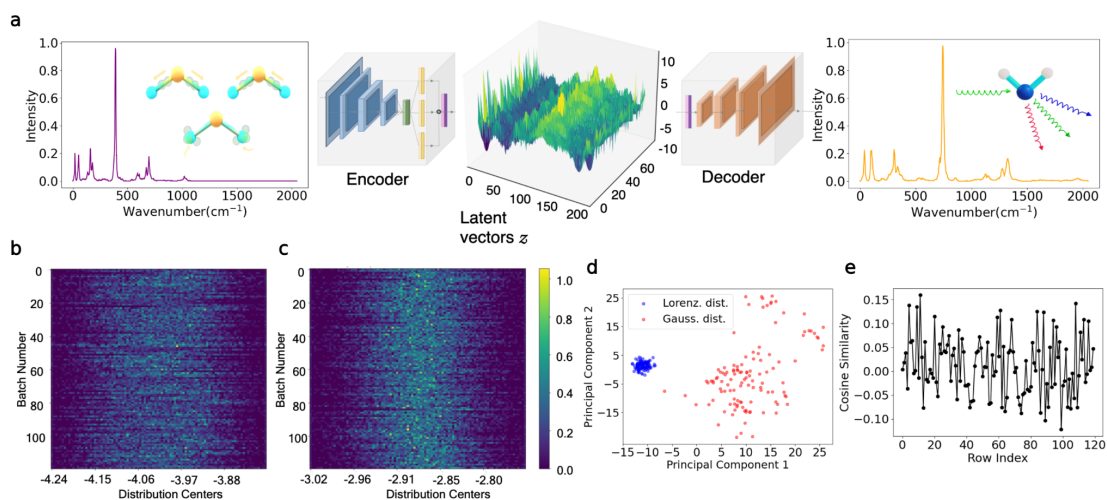

Fig.4 SpectroGen provides physical prior-informed spectrum deconvolution and generation. (a) The probabilistic encoder represents the input spectra with latent vectors $z$ with the guidance of physical priors. The probabilistic decoder learns the distribution from $z$ to map the generated spectrum. Even with the same input, different physical priors will result in discrepancies in latent vectors (see the detailed network parameter in Supplemental method 6). (b) Visualization of latent vectors of IR-Raman transfer with Gaussian distribution prior and (c) Lorentzian distribution prior. Latent vectors show differences in width, distribution centers, and values for each training epoch between Gaussian and Lorentzian distribution prior-guided experiments. (d) Principal component analysis for latent vector between Gaussian distribution prior and Lorentzian distribution prior. Gaussian prior led to a dispersed distribution in principal components 1 and 2 for the IR-to-Raman transformation, compared to a concentrated distribution with the same input data for the Lorentzian prior guided SpectroGen. (e) Cosine similarity with Gaussian and Lorentzian distribution prior. Both positive and negative cosine similarities indicate significant differences in the latent features in terms of their magnitudes and directions.

## Discussion

It has long been recognized that the analysis of interactions between matter and electromagnetic waves illuminates structure, property and function across broad fields such as biology, chemistry, and material sciences. Using the latest advancements in generative deep learning we demonstrate effective cross modality spectral transfer with precise multi-dimensional molecular, structural and other material property representation. SpectroGen, our physical prior-informed deep generative model, achieves state-of-the-art performance in cross-domain spectral transformation with generated spectra showing 99% average correlation and 0.01 RMSE compared to experimentally obtained spectra. Furthermore, transformation testing of multiple spectral modalities provides compelling evidence of SpectroGen's strong generalization capabilities. Experimental results indicate that, under the premise of objective, spectrum-based physical priors, we can accurately generate spectral data from another completely different spectroscopy modality. Although we cannot explicitly delineate the specific relationship between the latent features learned from the spectrum shape and the material structural property it represents, our results demonstrate comparable peak ratio, FWHM, and AUC. Notably, we have statistically significant improvement in SNR compared to experimentally generated spectra which led to a competitive classification test accuracy and 20% higher training performance.

This first of its kind demonstration promises spectroscopy implementation without the need for physical instrumentation, which is key as these modalities are often expensive or inaccessible. In addition, our approach is key for research work where sample-specific experimental challenges such as active specimens or in vivo biological samples, impose considerable limitations on spectral acquisition. SpectroGen effectively revolutionizes the reach of spectroscopy based analysis across disciplines and research areas. Our investigation, though modest, serves as a clear and direct representation of how computational technologies can synergize with principles from physics, chemistry, and biology to surmount objective barriers and drive the intelligent evolution of traditional high-precision methodologies. It is worthy to highlight that this approach not only enhances existing technologies but also could assist in pioneering novel spectroscopic methods, revealing previously uncharacterized material properties.

## Data availability
Data used in this research are all from open-source dataset RRUFF[24].

## Code availability
Code of this research is available at https://github.com/ymzhu19eee/Raman-generation

## Acknowledgements
This research was funded by the Massachusetts Institute of Technology, specifically through Prof. Loza F. Tadesse's startup package (1578169) from the Department of Mechanical Engineering. We thank Sujan Manna, Dr. Jeong-Hee

(Jenn) Kim and Dr. Jia Dong for their valuable suggestions and comments during the manuscript writing process.

**Author contributions**
Y.M.Z. proposed the idea, designed the model, wrote the code, conducted the experiments, wrote the manuscript, revised the manuscript. L.F.T. proposed the idea, revised the model, revised the manuscript, acquired the funding resource.

**Competing interests**
The authors declare no competing interests.